\pdfoutput=1

\documentclass[11pt]{article}

\usepackage{authblk}

\usepackage{acl}

\usepackage{times}
\usepackage{latexsym}

\usepackage[T1]{fontenc}

\usepackage[utf8]{inputenc}

\usepackage{microtype}

\usepackage{graphicx}
\usepackage{fontawesome}
\usepackage{bm}

\title{Detecting Narrative Elements in Informational Text}

\author[1]{Effi Levi}
\author[2]{Guy Mor}
\author[2,3]{Tamir Sheafer}
\author[2]{Shaul R. Shenhav}
\affil[1]{Institute of Computer Science, The Hebrew University of Jerusalem}
\affil[2]{Department of Political Science, The Hebrew University of Jerusalem}
\affil[3]{Department of Communication and Journalism, The Hebrew University of Jerusalem}
\affil[ ]{{\tt efle@cs.huji.ac.il}} 
\affil[ ]{{\tt \{guy.mor$|$tamir.sheafer$|$shaul.shenhav\}@mail.huji.ac.il}}

\begin{document}
\maketitle
\begin{abstract}
Automatic extraction of narrative elements from text, 
combining narrative theories with computational models,
has been receiving increasing attention over the last few years. Previous works have utilized the oral narrative theory by Labov and Waletzky to identify various narrative elements in personal stories texts. Instead, we direct our focus to informational texts, specifically news stories. 

We introduce NEAT (Narrative Elements AnnoTation) -- a novel NLP task for detecting narrative elements in raw text.
For this purpose, we designed a new multi-label narrative annotation scheme, better suited for informational text (e.g. news media), by adapting elements from the narrative theory of Labov and Waletzky (\texttt{Complication} and \texttt{Resolution}) and adding a new narrative element of our own (\texttt{Success}). We then used this scheme to annotate a new dataset of 2,209 sentences, compiled from 46 news articles from various category domains\footnote{\url{https://github.com/efle/NEAT}}.
We trained a number of supervised models in several different setups over the annotated dataset to identify the different narrative elements, achieving an average $F_1$ score of up to 0.77. The results demonstrate the holistic nature of our annotation scheme as well as its robustness to domain category.

\end{abstract}

\section{Introduction}
\label{sec:intro}

Automatic extraction of narrative elements from texts is a multidisciplinary field of research, 
combining narrative theories with computational models,
which has been receiving increasing attention over the last few years. Examples include modeling narrative structures for story generation \cite{gervas2006narrative}, using unsupervised methods to detect narrative event chains \cite{chambers-jurafsky-2008-unsupervised} 
and detecting content zones \cite{baiamonte2016annotating} in news articles, using semantic features to detect \textit{narreme} boundaries in fictitious prose \cite{delmonte-marchesini-2017-semantically}, identifying turning points in movie plots \cite{papalampidi2019movie} and using temporal word embeddings to analyze the evolution of characters in the context of a narrative plot \cite{volpetti2020temporal}.

A recent and more specific line of work focuses on using the theory laid out by \citet{labov1967narrative} and later refined by \citet{labov2013language} to characterize narrative elements in personal experience texts.
\citet{swanson-etal-2014-identifying} relied on \citet{labov1967narrative} to annotate a corpus of 50 personal stories from weblogs posts, and tested several models over hand-crafted features to classify clauses into three narrative clause types: \textit{orientation}, \textit{evaluation} and \textit{action}.
\citet{ouyang-mckeown-2014-towards} constructed a corpus from 20 oral narratives of personal experience collected by \citet{labov2013language}, and utilized logistic regression over hand-crafted features to detect instances of \textit{complicating actions}. 
More recently, \citet{li2017annotating} utilized a combination of ideas from \citet{labov1967narrative} and \citet{freytag1894technik} to annotate a collection of short stories, and \citet{saldias2020exploring} used convolutional neural networks (CNNs) to classify clauses from spoken personal texts into the same three narrative clause types as \citet{swanson-etal-2014-identifying}.

While these works concentrated their effort on narrative analysis of personal experience texts, we direct our focus to detecting narrative patterns 
in informational texts, such as news stories. The social impact of news stories distributed by the media and their role in creating and shaping of public opinion incentivized our efforts to adapt narrative analysis approaches to this domain. To the best of our knowledge, this is the first attempt to automatically detect narrative elements based on \citet{labov1967narrative} and later works by \citet{labov1972language, labov2013language} in news articles. 

In this work, we introduce NEAT (Narrative Elements AnnoTation) -- a novel NLP task for detecting narrative elements in raw text. For this purpose, we adapted two elements from the narrative theory presented in \citet{labov1967narrative, labov1972language, labov2013language}, namely {\tt Complication} and {\tt Resolution}, while adding a new narrative element, {\tt Success}, to create a new multi-label narrative annotation scheme.
This scheme was designed with two main objectives in mind. First, capturing elements oriented towards discourse structure, rather than semantic content. Second, possessing the flexibility required to capture narrative characteristics within a wide variety of text types, specifically informational text (as opposed to personal experience), and not only literary and well-structured stories. 
We used this scheme to annotate a newly-constructed dataset of 2,209 sentences, compiled from 46 English news articles; each sentence was tagged with a subset of the three narrative elements (or, in some cases, none of them), thus defining a novel multi-label classification task. 
 
We explored two different approaches towards solving our new task: splitting into three unrelated binary classification tasks ({\tt Complication}, {\tt Resolution} and {\tt Success}), and jointly learning the three narrative categories as a multi-label classification task. We experimented with three supervised models, each based on fine-tuning a different pre-trained language model: BERT \cite{devlin2018bert}, RoBERTa \cite{liu2019roberta} and DistilBERT \cite{sanh2020distilbert}, achieving an average $F_1$ score of up to 0.77. An analysis of the results indicates that our narrative categories are strongly connected and form a coherent narrative scheme which is more than just the sum of its parts. Additional experimentation with cross-domain classification demonstrates the task's robustness to domain category, suggesting that our annotation scheme is more grounded in discourse characteristics rather than semantic context.

The remainder of this paper is organized as follows: Section \ref{sec:narrativeanalysis} gives a theoretical background and describes the adjustments we have made to the scheme in \citet{labov2013language} in order to adapt it to informational text. Section \ref{sec:dataset} provides a complete description of the dataset and of the processes and methodologies which were used to construct and annotate it, along with a short analysis and some examples for annotated sentences. Section \ref{sec:experiments} describes the experiments conducted on the dataset, and Section \ref{sec:results} provides an analysis and a discussion of the results. Finally, Section \ref{sec:conclusion} contains a summary of our contributions as well as several potential directions for future work.

\section{Narrative Analysis}
\label{sec:narrativeanalysis}
\subsection{Background}

Ever since the emergence of formalism and structuralistic literary criticism \cite{propp1968morphology}
and throughout the development of narratology \cite{genette1980narrative, fludernik2009introduction, chatman1978existents, rimmon2003narrative},
narrative structure has been the focus of extensive theoretical and empirical research. While most of these studies were conducted in the context of literary analysis, the interest in narrative structures 
has made inroads into social sciences \cite{shenhav2015analyzing}. The classical work by \citet{labov1967narrative} on oral narratives, as well as later works \cite{labov1972language, labov2013language}, signify this stream of research by providing a schema for an overall structure of narratives, according to which a narrative construction encompasses the following building blocks  \cite{labov1972language, labov2013language}:
\textit{abstract} (what is the narrative about), \textit{orientation} (information on the time, the place, the persons and the behavior involved),  \textit{complicating action} (or simply complication; the forward progression of narrative clauses), \textit{evaluation} (establishing the narrative's "point"), \textit{resolution} (what finally happened), and \textit{coda} (bringing the time of reference back to the present time of narration).
These building blocks provide useful and influential guidelines for oral narratives analysis.

\subsection{Adaptation}
\label{subsec:adaptation}

\begin{table*}[!ht]
\centering
\begin{tabular}{lccc}
\hline 
& \textbf{Complication} & \textbf{Resolution} & \textbf{Success} \\ \hline
\# Sentences & 1,092 & 541 & 312 \\
Proportion in Dataset & 49\% & 24\% & 14\% \\
\hline
\end{tabular}
\caption{\label{tab:datasetoverview} Overview of the NEAT dataset. Note that the categories are not mutually exclusive, due to the multi-labeled nature of the annotation scheme.}
\end{table*}

Despite the substantial influence of \citet{labov1967narrative} and \citet{labov2013language}, scholars in the field of communication have noticed that this overall structure does not necessarily comply with the form of informational text, such as news stories \cite{thornborrow2004storying, van1988news},
and consequently proposed modified narrative structures \cite{thornborrow2004storying}. Unlike well-tailored narrative texts, such as personal experience texts,
narrativity in informational text is somewhat more challenging as it does not necessarily follow conventional or predefined genre-related structures. This requires a flexible coding scheme, unconstrained by a specific type of text. Instead, it should be open to a wide range of text types (such as informational text), and allow the presence of micro stories, encompassing any combination of all narrative categories even at the sentence level. 
We set to accomplish that via two objectives: first, 
formalizing narrative categories which are oriented towards discourse structure, rather than semantic context. Second, defining our task as a multi-labeled one, to allow the flexibility required to capture sentence-level narrative characteristics.
A special consideration was given to the variety of contents, forms and writing styles typical for media texts. For example, we required a coding scheme that would fit laconic or problem-driven short reports (too short for full-fledged “Labovian” narrative style), as well as complicated texts with multiple story-lines moving from one story to another. We addressed this challenge by focusing on two of Labov’s six elements - \textit{complicating action} and \textit{resolution}, considered 
to be the most fundamental and relevant for informational text analysis \cite{labov2013language}. 
There are several reasons for our focus on these particular elements: first, it goes in line with the understanding that worth-telling stories usually consist of protagonists facing and resolving problematic experiences \cite{eggins2005analysing}.
Moreover, these elements resonate with what is considered by \citet{entman2004projections} to be the most important Framing Functions - problem definition and remedy. 

In order to adapt the original \textit{complicating action} and \textit{resolution} categories to informational content, we designed our annotation scheme as follows. \textit{Complicating action} -- hence, \texttt{Complication} -- was defined in our narrative scheme as an event, series of events or situation, that point at problems or tensions. \texttt{Resolution} refers to the way the story is resolved or to the release of the tension. 
An improvement from -- or a manner of coping with -- an existing or a hypothetical situation was also considered to be
a \texttt{Resolution}. This choice was made in order to follow 
the often tentative or speculative notion of future resolutions in news stories \cite{thornborrow2004storying, bell1991language}. We have therefore included in this category any temporary or partial resolutions. The transitional characteristic of the \texttt{Resolution} motivated us to add a new category defined as \texttt{Success}. Unlike 
\texttt{Resolution}, which refers, implicitly or explicitly, to a prior situation, this category was designed to capture any description or indication of an achievement or a desirable outcome. 

\section{The Dataset}
\label{sec:dataset}

\subsection{Pilot Study}
\label{subsec:pilot}
We started by conducting a pilot study, for the purpose of formalizing an annotation scheme and training our annotators. For this study, 
sample sentences were gathered from print news articles, published between 1995 and 2017 and collected via {\it LexisNexis}. These were used to refine
the annotation scheme described in Section \ref{subsec:adaptation}, as well as perform extensive training for our annotators. 

Following the conclusion of the pilot study, we used the 
sentences which were collected and manually annotated during the pilot 
to train a multi-label classifier, later used to provide labeled candidates for the annotators during the annotation stage of the NEAT dataset, in order to optimize annotation rate and accuracy. The pilot samples were then discarded.

\subsection{News Articles}

The news articles for the dataset were sampled from leading news websites in the English language, all published between 2017 and 2020. The result is a corpus of 2,209 sentences taken from 46 news articles, with an average of 48 sentences per article ($\sigma^2=39.44$), and an average of 20.2 tokens per sentence ($\sigma^2=11.2$). 
The articles are semantically diverse, as they were sampled from a wide array of domain categories.

\subsection{Preprocessing}

The news articles' content was extracted using \href{http://diffbot.com}{diffbot}. The texts were scraped and split into sentences using the Punkt unsupervised sentence segmenter \cite{kiss-strunk-2006-unsupervised}. Remaining segmentation errors were manually corrected. 

\subsection{Annotation}

\subsubsection{Guidelines}
\label{subsubsec:guidelines}

Following the pilot study (Section \ref{subsec:pilot}), a code book containing annotation guidelines was produced. For each of the three categories in the annotation scheme -- {\tt Complication}, {\tt Resolution} and {\tt Success} -- the guidelines provide: 
\begin{itemize}
    \itemsep0em
    \item A general explanation of the category
    \item A list of well-defined criteria for identifying the category
    \item Select examples of sentences labeled with the category
\end{itemize}

\subsubsection{Process}

We employed a three-annotator setup for annotating the collected sentences. First, the pilot stage model (Section \ref{subsec:pilot}) was used to produce annotation suggestions for each of the sentences in the corpus. Each sentence was then separately annotated by two trained annotators according to the guidelines described in Section \ref{subsubsec:guidelines}. Each annotator had the choice to either accept the suggested annotation or to change it by adding or removing any of the suggested labels. Disagreements were later decided by a third expert annotator.
Table \ref{tab:intercoderreliability} reports inter-coder reliability scores for each of the three categories, averaged across pairs of annotators: pairwise percent agreement (PPA), and Cohen's Kappa coefficient, accounting for chance agreement \cite{artstein-poesio-2008-survey}. 
Article-level domain categories (Table \ref{tab:domains}) were initially assigned according to the news section from which the articles were taken, and later verified by two annotators. 

\begin{table}[!ht]
\centering
\begin{tabular}{lcc}
\hline
& \textbf{PPA} (avg.) & $\bm{\kappa}$ (avg.) \\
\hline
Complication & 89.6\% & 0.79 \\
Resolution & 87.5\% & 0.65 \\
Success & 92.8\% & 0.7 \\
\hline
\end{tabular}
\caption{\label{tab:intercoderreliability} Inter-coder reliability: average pairwise percent agreement (PPA) \& average Cohen's Kappa ($\kappa$)}
\end{table}

\subsection{Analysis}
\label{subsec:analysis}

Narrative categories vary significantly in their prevalence in the corpus; their respective proportions in the dataset are given in Table \ref{tab:datasetoverview}. The categories are unevenly distributed: {\tt Complication} is significantly more frequent than {\tt Resolution} and {\tt Success}. This was to be expected, considering the known biases of "newsworthiness" towards problems, crises and scandals \cite{esser2016negativity}, and due to the fact that in news media, resolutions often follow reported complications.

Interestingly, the distribution over narrative categories varies significantly between the different category domains (see Table \ref{tab:domains}). Most domains contain many more \texttt{Complications} than \texttt{Resolutions} or \texttt{Successes}, which is consistent with the distribution in the complete dataset (Table \ref{tab:datasetoverview}); the ``Crime'' domain is an extreme example, with a very small number of \texttt{Resolutions} and no \texttt{Successes} at all. However, some domains exhibit a completely different distribution. For example, the ``Travel'' and the ``Science \& Technology'' domains possess a relatively uniform distribution over the three narrative categories. 
The ``Sports'' domain contains a similar number of \texttt{Complications} and \texttt{Successes}, with a smaller number of \texttt{Resolutions}.

\begin{table*}
\centering
\begin{tabular}{ccccc}
\hline 
\textbf{Domain Category} & \textbf{\# Sentences} & \textbf{Complication} & \textbf{Resolution} & \textbf{Success} \\ 
\hline
Economy \& Job Market & 221 & 99 & 71 & 25 \\
Politics & 860 & 514 & 220 & 89 \\
Health & 66 & 38 & 13 & 10 \\
Travel & 200 & 47 & 46 & 42 \\
Arts \& Culture & 345 & 166 & 63 & 49 \\
Crime & 28 & 25 & 3 & 0 \\
Accidents \& Disasters & 12 & 9 & 4 & 2 \\
Welfare & 47 & 30 & 11 & 1 \\
Sports & 115 & 51 & 21 & 47 \\
Science \& Technology & 223 & 53 & 60 & 40 \\
Immigration & 92 & 60 & 29 & 7 \\
\hline
\end{tabular}
\caption{\label{tab:domains} Domain category distribution in the NEAT dataset}
\end{table*}

Table \ref{tab:label-correlations-table} reports pairwise Pearson correlations between the categories. The {\tt Complication} and {\tt Resolution} categories
are 
completely uncorrelated ($r=0.016$). A minor negative correlation was found between {\tt Complication} and {\tt Success} ($r=-0.234$), and a minor positive 
one 
was found between {\tt Resolution} and {\tt Success} ($r=0.228$). These minor correlations -- in our opinion -- indicate that the \texttt{Success} category does indeed bring added value to our narrative scheme.  

\begin{table}
\centering
\begin{tabular}{lccc}
\hline & \textbf{Comp.} & \textbf{Res.} & \textbf{Suc.} \\ \hline
\textbf{Comp}. & 1 & & \\
\textbf{Res.} & 0.016 & 1 & \\
\textbf{Suc.} & -0.234 & 0.228 & 1 \\
\hline
\end{tabular}
\caption{\label{tab:label-correlations-table} Inter-category Pearson correlations}
\end{table}

\begin{table}
\centering
\begin{tabular}{lc}
\hline 
\textbf{Combination} & \textbf{\# Sentences} \\ \hline
Complication \& Resolution & 226 \\
Complication \& Success & 15 \\
Resolution \& Success & 103 \\
All Three Categories & 49 \\
\hline
\end{tabular}
\caption{\label{tab:category-combinations} Narrative category co-occurrences in the NEAT dataset}
\end{table} 

All the possible combinations of narrative categories appear in the dataset; Table \ref{tab:category-combinations} summarizes the occurrences of each of the possible category combinations. Examples of sentences annotated with various category combinations are given in 
Appendix \ref{app:examples}. 
There is, however, a significant variability to the frequency in which different combinations occur in the dataset. For example, the \texttt{Complication}-\texttt{Resolution} combination, designating a typical narrative tension-relief pattern \cite{shenhav2015analyzing}, is by far the most frequent one with 226 sentences. 
\texttt{Complication}-\texttt{Success}, on the other hand, is a very rare combination with only 15 sentences, embodying a far less trivial or common logic, where a success is accompanied by an unresolved problem. 

The fact that the dataset is assembled from full coherent news articles allows the analysis of a range of micro and macro stories in narrative texts. For example, an article in the dataset concerning the coronavirus outbreak in South Korea\footnote{\url{https://edition.cnn.com/2020/03/09/asia/south-korea-coronavirus-intl-hnk/index.html}} opens with a one-sentence summary, tagged with both \texttt{Complication} and \texttt{Resolution}:
\begin{itemize}
    \item[] "South Korea's top public health official hopes that the country has already gone through the worst of the novel coronavirus outbreak that has infected thousands inside the country." (\texttt{Complication}, \texttt{Resolution})
\end{itemize}

This problem-solution 
(in this case, hopeful solution) plot structure reappears in the 
article, 
this time 
detailed over a series of sentences: 
\begin{itemize}
    \itemsep0em 
    \item[] “More than 7,300 coronavirus infections have been confirmed throughout South Korea, killing more than 50." (\texttt{Complication})
    \item[] "It is one of the largest outbreaks outside mainland China, where the deadly virus was first identified.” (\texttt{Complication})
    \item[] “However, the number of new daily infections in South Korea has declined in recent days.” (\texttt{Complication}, \texttt{Resolution})
    \item[] “… while he believes the aggregate number of infections is high, he is confident in the job South Korea did to combat the virus' spread and would advise other governments…” (\texttt{Complication}, \texttt{Resolution})
    \item[] “The South Korean government has been among the most ambitious when it comes to providing the public with free and easy testing options." (\texttt{Success})
\end{itemize}

The sequence starts with two sentences tagged with \texttt{Complication}, followed by two 
sentences tagged with 
\texttt{Complication} and \texttt{Resolution}, and concludes with a sentence tagged with \texttt{Success},
demonstrating
a more gradual transition from problem through solution to success.

\section{Experiments}
\label{sec:experiments}

\subsection{Dataset Partition}
\label{subsec:partition}

We randomly divided the dataset into article-wise mutually-exclusive train, validation and test sets (details in Table \ref{tab:datasetdivision}), while keeping the distribution over the three narrative categories in each of the sets as similar as possible to the one in the complete dataset. 
The train set was used to train a supervised model for the task; the validation set was used to select the best model configuration during the training phase by tuning the model's hyper-parameters (see Section \ref{subsec:models} for details), and the test set was used to evaluate the chosen model and produce the results reported in Section \ref{sec:results}.

\begin{table}[hbt!]
\centering
\begin{tabular}{lccc}
\hline & \textbf{\# Sentences (Articles)} & \textbf{Ratio} \\ \hline
Train & 1,767 (37) & 80\% \\
Validation & 222 (5) & 10\% \\
Test & 220 (4) & 10\% \\
\hline
\end{tabular}
\caption{\label{tab:datasetdivision} Train, validation \& test set statistics}
\end{table}

\subsection{Task Definition}
\label{subsec:taskdef}

We explored two different approaches for solving the task: (1) addressing each of the three narrative categories as a separate classification task, and (2) treating the task as multi-label classification with three labels (one for each narrative category).

\subsubsection{Separate Classification Tasks}
\label{subsubsec:seperatetasks}
In this approach, we defined a separate binary classification task for each of the narrative categories: {\tt Complication}, {\tt Resolution} and {\tt Success}. For each such task, we trained a dedicated supervised model (further details given in Section \ref{subsec:models}), specifically optimized for the learned category. However, any potential information stemming from inter-correlations between the different categories was ignored, effectively treating them as three unrelated tagging schemes. 

\subsubsection{Multi-Label Classification}
\label{subsubsec:multilabeltask}
Here, the task was treated as a three-way multi-label classification problem (each sentence may contain any combination of the three narrative categories), thus taking advantage of inter-correlations between the three narrative categories to better learn them as part of a coherent narrative scheme. We trained and optimized a single multi-label model to jointly predict the three categories (further details given in Section \ref{subsec:models}).

\subsection{System Architecture}
\label{subsec:models}

We employed the method of fine-tuning a pre-trained language model for our task. In each experimental setup, we chose a pre-trained language model as a backbone, applied a multilayer perceptron (MLP) classifier on top of it, and fine-tuned the entire model over the train set.

\subsubsection{Backbone}
\label{subsubsec:backbone}

We experimented with three state-of-the-art transformer-based language models as the backbone for our inference model, using pre-trained weights from the {\it transformers} python package \cite{wolf2019huggingfaces}.

\textbf{BERT.} Following common practice, we first utilized the base-sized BERT \cite{devlin2018bert} as the backbone model.

\textbf{RoBERTa.} This BERT variant was developed by training the original BERT model with altered design choices and training techniques, and has been recently shown to produce better results on various NLP tasks \cite{liu2019roberta}. We used the base-sized RoBERTa as the backbone model.

\textbf{DistilBERT.} A recent body of work has focused on developing ``lighter'' transformer-based language models which allow for faster fine-tuning for downstream NLP tasks 
\cite{sanh2020distilbert, lan2020albert}. In order to assess robustness to a decrease in the model's size, we also experimented with DistilBERT \cite{sanh2020distilbert}, which follows the same basic architecture as BERT but consists of 66M parameters (as opposed to 110M in BERT), as the backbone model.

\subsubsection{Classifier}

In order to fine-tune the backbone language model, we appended a multilayer perceptron (MLP) over the output of the language model. The MLP consisted of one hidden layer (increasing the number of hidden layers produced no improvement in performance), the size of which was optimized as a hyper-parameter. In the case of a single binary classification task (Section \ref{subsubsec:seperatetasks}), the output layer consisted of a single sigmoid output, while in the case of a multi-labeled task, it consisted of three sigmoid outputs, one for each narrative category.

\subsubsection{Training Procedure}
\label{subsubsec:training}

All models were optimized using the AdamW algorithm \cite{loshchilov2017decoupled} and the binary cross entropy loss function. Positive weighting was used in order to compensate for class imbalance (evident in Table \ref{tab:datasetoverview}). Hyper-parameters -- batch size, learning rate and MLP hidden layer size -- were chosen via a standard grid search (see Appendix \ref{app:hyperparameters} for more details). For each configuration of task definition, backbone model and hyper-parameters, the model was evaluated over the validation set after every epoch of training, and the best-performing checkpoint was tested on the test set to produce the results reported in Section \ref{sec:results}.

\begin{table*}[ht]
\resizebox{\textwidth}{!}{
\centering
\begin{tabular}{cc|ccccccccc|ccc}
\hline 
& & \multicolumn{3}{c}{\textbf{Complication}} & \multicolumn{3}{c}{\textbf{Resolution}} & \multicolumn{3}{c}{\textbf{Success}} & \multicolumn{3}{|c}{\textbf{Average}} \\
Task & Backbone & P & R & $F_1$ & P & R & $F_1$ & P & R & $F_1$ & P & R & $F_1$ \\
\hline 
Separate & BERT & 0.89 & 0.87 & 0.88 & 0.67 & 0.40 & 0.50 & 0.33 & 0.72 & 0.45 & 0.63 & 0.66 & 0.61 \\
Separate & RoBERTa & \textbf{0.92} & 0.85 & 0.88 & 0.69 & 0.56 & 0.62 & \textbf{0.83} & 0.56 & 0.67 & \textbf{0.81} & 0.66 & 0.72 \\
Separate & DistilBERT & 0.87 & 0.84 & 0.85 & 0.69 & 0.40 & 0.51 & 0.53 & 0.44 & 0.48 & 0.70 & 0.56 & 0.61 \\
\hline
M.Label & BERT & 0.88 & 0.89 & 0.88 & 0.72 & 0.60 & 0.65 & 0.53 & 0.50 & 0.51 & 0.71 & 0.66 & 0.68 \\
M.Label & RoBERTa & 0.90 & \textbf{0.90} & \textbf{0.90} & \textbf{0.76} & \textbf{0.67} & \textbf{0.71} & 0.64 & \textbf{0.78} & \textbf{0.70} & 0.77 & \textbf{0.78} & \textbf{0.77} \\
M.Label & DistilBERT & 0.86 & 0.87 & 0.86 & 0.73 & 0.53 & 0.61 & 0.75 & 0.50 & 0.60 & 0.77 & 0.63 & 0.69 \\
\hline
\end{tabular}}
\caption{\label{tab:results} Test set precision (P), recall (R) and $F_1$ scores, for every combination of task definition (a separate task for each narrative category / a multi-labeled task) and backbone model. See Sections \ref{subsec:taskdef} and \ref{subsubsec:backbone} for details.}
\end{table*}

\subsection{Cross-Domain Classification}
\label{subsec:cross-domain}

Given the semantic diversity in the dataset, as well as the variability in distribution over the narrative categories between the various domains (Table \ref{tab:domains}), we wished to assess the domain category's effect on learning our narrative scheme. For this purpose, we experimented with a cross-domain classification setup. For each of the eleven category domains, we concatenated the sentences from all other domains into a train set, which was then used to train a classification model. The training process was done using the configuration of the best-performing model from the previous stage (described in Sections \ref{subsec:partition}--\ref{subsec:models}), including task definition, backbone model and hyper-parameters (i.e. no hyper-parameter tuning was performed in this setup). The trained model was then evaluated on the test set.

\section{Results \& Discussion}
\label{sec:results}

Results are reported in Table \ref{tab:results}. For each task definition and choice of backbone model, we report the precision, recall and $F_1$ score for each of the three narrative categories, as well as their average, over the test set. 

\begin{table*}[ht]
\resizebox{\textwidth}{!}{
\centering
\begin{tabular}{c|ccccccccc|ccc}
\hline 
& \multicolumn{3}{c}{\textbf{Complication}} & \multicolumn{3}{c}{\textbf{Resolution}} & \multicolumn{3}{c}{\textbf{Success}} & \multicolumn{3}{|c}{\textbf{Average}} \\
Domain Category & P & R & $F_1$ & P & R & $F_1$ & P & R & $F_1$ & P & R & $F_1$ \\
\hline
Accidents \& Disasters & 1.00 & 0.89 & 0.94 & 1.00 & 0.50 & 0.67 & 1.00 & 0.50 & 0.67 & 1.00 & 0.63 & 0.76 \\
Arts \& Culture & 0.83 & 0.84 & 0.84 & 0.67 & 0.75 & 0.71 & 0.51 & 0.78 & 0.62 & 0.67 & 0.79 & 0.72 \\
Crime & 1.00 & 0.92 & 0.96 & 0.67 & 0.67 & 0.67 & - & - & - & 0.83 & 0.79 & 0.81 \\
Economy \& Job Market & 0.86 & 0.78 & 0.81 & 0.77 & 0.68 & 0.72 & 0.50 & 0.84 & 0.63 & 0.71 & 0.76 & 0.72 \\
Health & 0.97 & 0.82 & 0.89 & 0.63 & 0.77 & 0.69 & 0.67 & 0.80 & 0.73 & 0.75 & 0.80 & 0.77 \\
Immigration & 0.86 & 0.93 & 0.90 & 0.80 & 0.83 & 0.81 & 0.71 & 0.71 & 0.71 & 0.79 & 0.83 & 0.81 \\
Politics & 0.83 & 0.95 & 0.89 & 0.58 & 0.86 & 0.69 & 0.60 & 0.65 & 0.63 & 0.67 & 0.82 & 0.74 \\
Science \& Technology & 0.75 & 0.83 & 0.79 & 0.66 & 0.77 & 0.71 & 0.64 & 0.88 & 0.74 & 0.68 & 0.82 & 0.74 \\
Sports & 0.87 & 0.65 & 0.74 & 0.53 & 0.76 & 0.63 & 0.84 & 0.81 & 0.83 & 0.75 & 0.74 & 0.73 \\
Travel & 0.89 & 0.87 & 0.88 & 0.66 & 0.89 & 0.76 & 0.69 & 0.83 & 0.75 & 0.75 & 0.87 & 0.80 \\
Welfare & 1.00 & 0.97 & 0.98 & 0.80 & 0.73 & 0.76 & 0.33 & 1.00 & 0.50 & 0.71 & 0.90 & 0.75 \\
\hline
\textbf{Average} & \textbf{0.90} & \textbf{0.86} & \textbf{0.87} & \textbf{0.71} & \textbf{0.74} & \textbf{0.71} & \textbf{0.65} & \textbf{0.78} & \textbf{0.68} & \textbf{0.76} & \textbf{0.79} & \textbf{0.76} \\
\hline
\end{tabular}}
\caption{\label{tab:cross-domain} Precision (P), recall (R) and $F_1$ scores for cross-domain classification. See Section \ref{subsec:cross-domain} for details.}
\end{table*}

First, we observe that addressing the task as a multi-labeled one proved to be a better strategy than learning each narrative category separately. This is evident across backbone models as well as across narrative categories; for each backbone model, the multi-label model produced a higher $F_1$ score for each and every one of the narrative categories. 
This is a clear indication that these categories are substantially connected as they constitute intertwining elements in an underlying story. Therefore, the three categories form a coherent narrative scheme that is more than just the sum of its parts. 

Interestingly, while this effect is relatively small for \texttt{Complication} ($F_1$ increased by 0.00--0.02), it is much more prominent for \texttt{Resolution} ($F_1$ increased by 0.09--0.15) and \texttt{Success} ($F_1$ increased by 0.03--0.12), meaning that incorporating all three categories into one coherent scheme contributes mostly to learning the \texttt{Resolution} and \texttt{Success} categories. 
This suggests that perhaps the narrative properties of the \texttt{Complication} category make it more independent and self-contained than the other two categories.
\texttt{Resolution} and \texttt{Success}, on the other hand, are more relative in nature, and seem to be anchored, implicitly or explicitly, by a prior situation or condition. 

Among the three narrative categories, \texttt{Complication} gained the highest $F_1$ scores by all the models, ranging between 0.85 and 0.90. The models were less successful predicting \texttt{Resolution}, with $F_1$ scores ranging between 0.50 and 0.71, and \texttt{Success}, with $F_1$ scores ranging between 0.45 and 0.70. 
This is consistent with the proportion of instances belonging to each category in the dataset (see Table \ref{tab:datasetoverview}), which may provide a possible explanation for this observation. However, the fact that positive weighting was used in the training process (Section \ref{subsubsec:training}) to counter class imbalance, motivates us to search for another explanation. Defined by \citet{labov1967narrative}'s overall structure of narratives as ``the main body of narrative clause'', \texttt{Complication} may just be an easier narrative category to learn.

Comparing different backbone models, the DistilBERT-based model performed similarly to the BERT-based one -- an average $F_1$ score of 0.61 for both in the separate-task setting, and 0.69 compared to 0.68 in the multi-label setting -- suggesting that the task is fairly robust to a decrease in the backbone model's size. However, RoBERTa significantly outperformed the other two language models as the backbone in both settings -- an average $F_1$ score of 0.71 compared to 0.61 in the separate-task setting, and an average $F_1$ score of 0.77 compared to 0.68 and 0.69 in the multi-label setting. We also note that the difference in performance between  \texttt{Complication} and the other two categories is less extreme in the RoBERTa-based models compared to the other backbone models. 

\subsection{Cross-Domain Classification}

The best-performing configuration (a multi-label classifier based on the RoBERTa language model) was used to perform the cross-domain classification experiment. As stated in Section \ref{subsec:cross-domain}, hyper-parameters were fixed to the values obtained in the train-validation-test setup. Results are presented in Table \ref{tab:cross-domain}. Averaged over domain categories, they are virtually identical to the results obtained on the test set (reported in Table \ref{tab:results}), with a precision, recall and $F_1$ score of 0.76, 0.79 and 0.76 compared to 0.77, 0.78 and 0.77 (respectively). In our opinion, this demonstrated invariance to domain category is a strong indication that our narrative elements are more grounded in discourse characteristics rather than in the semantic field.

\section{Conclusion}
\label{sec:conclusion}

We introduced NEAT (Narrative Elements AnnoTation) - a novel NLP task for detecting narrative elements 
in raw text. For this purpose, we designed a new flexible multi-label narrative annotation scheme, specifically suited for informational text, by adapting two elements from the theory introduced in \citet{labov1967narrative, labov1972language, labov2013language} -- \texttt{Complication} and \texttt{Resolution} -- and adding a new element -- \texttt{Success}. The scheme was used to annotate a new dataset of 2,209 sentences, compiled from 46 articles, which were collected from news websites.

We explored two alternate settings for solving this task - one in which each narrative category was treated as a separate classification task, and another in which the entire task was addressed as multi-label classification. 
In each of these setups, we experimented with fine-tuning three different language models,
achieving an average $F_1$ score of up to 0.77 on the test set, and showcasing the potential of supervised-learning methods in detecting the narrative information encoded into our scheme. 
The multi-label setting consistently provided significantly better results across all models and narrative categories, demonstrating that our narrative categories are strongly connected and form a coherent narrative scheme which is more than just the sum of its parts. Additional cross-domain classification results demonstrate the task's invariance to domain category, suggesting that our annotation scheme is more grounded in discourse characteristics rather than semantic context.

We are currently engaged in an ongoing effort for improving the annotation quality of the dataset and increasing its size. In addition, we have several interesting directions for future work. The first one, which we are currently pursuing, includes enriching the scheme with token-level annotation of the narrative elements, effectively converting the task from multi-label classification to a sequence prediction one.
Alternatively, we could introduce additional layers of information to encode more global narrative structures in the text, such as inter-sentence --  or even inter-article -- references between narratively-related elements (e.g., a \texttt{Resolution} referencing its inducing \texttt{Complication}).
Another potential direction is incorporating additional narrative elements into our annotation scheme. For example, the \textit{evaluation} element from \cite{labov2013language} may be beneficial in encoding additional information in the context of news media, such as the severity of a \texttt{Complication} or the `finality' of a \texttt{Resolution}. We could also add completely new narrative elements, tailored to capture specific informational aspects, such as actor-based elements identifying entities which are related to one or more of the currently defined narrative categories.

\section*{Acknowledgements}
This research was supported by the Israel Science Foundation (Grants No. 1400/14; 2315/18). 
We thank the three annotators -- Gal Ron, Vered Porzycki and Avishai Green -- for their dedicated work, as well as their insightful comments and suggestions during the annotation processes.

\bibliography{naacl2022}
\bibliographystyle{acl_natbib}

\clearpage

\appendix

\section*{Appendix}
\renewcommand{\thesubsection}{\Alph{subsection}}

\subsection{Sample Annotated Sentences}
\label{app:examples}

\begin{enumerate}
    \item How did some of the biggest brands in care delivery lose this much money?	(\texttt{Complication})
    \item Bleeding from the eyes and ears is also possible after use, IDPH said. (\texttt{Complication})
    \item His proposal to separate himself from his business would have him continue to own his company, with his sons in charge. (\texttt{Resolution})
    \item Instead, hospitals are pursuing strategies of market concentration. (\texttt{Resolution})
    \item With its centuries-old canals, vibrant historic center and flourishing art scene, Amsterdam takes pride in its cultural riches. (\texttt{Success})
    \item Mr. Trump chose to run for president, he won and is about to assume office as the most powerful man in the world. (\texttt{Success})
    \item Soon after, her administration announced a set of measures intended to curb misconduct. (\texttt{Complication, Resolution})
    \item Avant-gardists, the couple opened an art gallery in 1875 within the department store, offering artists who had been turned away by the Paris Salon -- the official art exhibition of the Academy of Fine Arts in Paris -- a home for their works and a large public audience, Burckhardt wrote. (\texttt{Complication, Resolution})
    \item Unlike Macy's, the well-known US department store which has been closing its doors around the country, Le Bon Marché isn't fighting to stay alive. (\texttt{Complication, Success})
    \item He defeated Bolt in a close semifinal heat, ending Bolt's 28-race winning streak in the 100 meters. (\texttt{Complication, Success})
    \item The Utah man's mother, Laurie Holt, thanked Mr. Trump and the lawmakers for her son's safe return, adding: "I also want to say thank you to President Maduro for releasing Josh and letting him to come home." (\texttt{Resolution, Success})
    \item And if Tony Hughes can keep his job without the weeks away, "in that aspect," he said, "it's going to make my life better." (\texttt{Resolution, Success})
    \item They were fortunate to escape to America and to make good lives here, but we lost family in Kristallnacht. (\texttt{Complication, Resolution, Success})
    \item He was vulnerable, no doubt, but in the past, with the spotlight at its brightest, he still found the speed and the will to remain the fastest man in the world. (\texttt{Complication, Resolution, Success})
\end{enumerate}

\subsection{Hyper-Parameters}
\label{app:hyperparameters}

Table \ref{tab:hyperparameters} lists the ranges of values which were used for hyper-parameter tuning during the experiments described in Section \ref{sec:experiments}.

\begin{table}[ht]
\centering
\begin{tabular}{lc}
\hline 
\textbf{Hyper-Parameter} & \textbf{Values} \\ \hline
Batch Size & 16, 32, 64 \\
Learning Rate & 1e-5, 2e-5, 3e-5, 4e-5, 5e-5 \\
MLP Hidden Size & 50, 100, 200, 300 \\
\hline
\end{tabular}
\caption{\label{tab:hyperparameters} Hyper-parameters value ranges}
\end{table} 

The best-performing model -- a multi-label classifier based on the RoBERTa language model -- was trained for 4 epochs with a batch size of 32, a learning rate of 4e-5 and an MLP classifier with a 100-node hidden layer.

\end{document}